\documentclass[letterpaper, 10 pt, conference]{ieeeconf}  

\IEEEoverridecommandlockouts                              

\usepackage{times} 
\usepackage{amsmath} 
\usepackage{amssymb}  

\usepackage{cite}
\usepackage{graphicx}
\usepackage{array}

\usepackage{lineno}
\usepackage{color}
\usepackage{xcolor}

\makeatletter
\let\NAT@parse\undefined
\makeatother
\usepackage{hyperref}

\usepackage{xspace}
\newcommand{\our}{\textsc{PALo}\xspace}

\title{\LARGE \bf
\our: Learning Posture-Aware Locomotion for Quadruped Robots
}

\author{Xiangyu Miao$^{1}$, Jun Sun$^{2*}$, Hang Lai$^{1}$, Xinpeng Di$^{2}$, Jiahang Cao$^{1}$, Yong Yu$^{1}$ and Weinan Zhang$^{1*}$
\thanks{$^{1}$Shanghai Jiao Tong University}%
\thanks{$^{2}$Shanghai Aerospace Control Technology Institute, China}%
\thanks{*Correspondence to: Jun Sun (sjlovedh@hotmail.com), Weinan Zhang (wnzhang@sjtu.edu.cn)}%
}

\begin{document}

\maketitle
\thispagestyle{empty}
\pagestyle{empty}

\begin{abstract}

With the rapid development of embodied intelligence, locomotion control of quadruped robots on complex terrains has become a research hotspot. Unlike traditional locomotion control approaches focusing solely on velocity tracking, we pursue to balance the agility and robustness of quadruped robots on diverse and complex terrains. To this end, we propose an end-to-end deep reinforcement learning framework for posture-aware locomotion named \our, which manages to handle simultaneous linear and angular velocity tracking and real-time adjustments of body height, pitch, and roll angles. 
In \our, the locomotion control problem is formulated as a partially observable Markov decision process, and an asymmetric actor-critic architecture is adopted to overcome the sim-to-real challenge. 
Further, by incorporating customized training curricula, \our achieves agile posture-aware locomotion control in simulated environments and successfully transfers to real-world settings without fine-tuning, allowing real-time control of the quadruped robot's locomotion and body posture across challenging terrains. Through in-depth experimental analysis, we identify the key components of \our that contribute to its performance, further validating the effectiveness of the proposed method. The results of this study provide new possibilities for the low-level locomotion control of quadruped robots in higher-dimensional command spaces and lay the foundation for future research on upper-level modules for embodied intelligence.

\end{abstract}

\section{INTRODUCTION}

Recent advancements in artificial intelligence have driven progress in embodied intelligence, unlocking new possibilities for quadruped robots \cite{nygaard2021real}. The motion intelligence of quadruped robots is typically divided into high-level motion planning and low-level motion control. Designing locomotion controllers remains a significant challenge due to the robots' high degrees of freedom (DOF) and the need to adapt to diverse structured and unstructured terrains. Traditional approaches rely on model-based control methods \cite{kang_animal_2021, yang_fast_2022, li_fastmimic_2023, rathod_model_2021}, where controllers are manually designed based on the robot's dynamics and control theory. However, these approaches involve complex pipelines and require extensive expert knowledge for accurate modeling and parameter tuning.

In contrast, deep reinforcement learning (RL) provides an end-to-end learning paradigm, enabling agents to directly map observations to actions through interaction with the environment, and reducing reliance on expert intervention \cite{mnih2013playing,sutton1998reinforcement}. Considering the low sampling efficiency and safety risks of directly training in the real world, a more practical alternative is to train the model in simulation and deploy it in the real world, also known as \emph{sim-to-real transfer} \cite{lai_sim--real_2023, zhao_sim--real_2020, tan_sim--real_2018}.

Typically, in quadrupedal locomotion, an RL policy is trained to follow a 3-dimensional (3D) velocity command, including linear velocities along the x- and y-axes and angular velocity around the z-axis in the robot's body frame, as shown in Fig.~\ref{fig:robot_simulation}. However, many real-world tasks require precise control of body posture in addition to velocity. For instance, in object transportation, the primary goal is to move from point A to point B, but maintaining object stability is equally important. Suppose a quadruped robot is tasked with delivering a cup of beverage, if it cannot adjust its body posture to prevent spillage, mere point-to-point locomotion would be insufficient, potentially leading to task failure. Similarly, in industrial inspections or terrain exploration, the robot may need to crouch to pass through confined spaces or adjust its orientation for better visibility. 
Despite these demands, most existing works focus solely on velocity tracking, neglecting explicit posture control \cite{smith_walk_2022, hwangbo_learning_2019, miki_learning_2022, agarwal_legged_2022, lai_sim--real_2023}. Some even penalize non-standard body orientations, prioritizing stability at the cost of agility \cite{ji_concurrent_2022, kumar_rma_2021, nahrendra_dreamwaq_2023}.

To enhance the controllability of quadruped robot locomotion, we propose an end-to-end deep reinforcement learning framework for \textbf{P}osture-\textbf{A}ware \textbf{Lo}comotion named \our, which enables adaptive control policies based on proprioceptive feedback and extended command inputs. \our allows the robot to execute agile, robust, and blind locomotion control under diverse body postures, master various motion skills, and traverse challenging terrains. Notably, the controller is trained entirely in simulation and transferred to real-world deployment without fine-tuning. In real-world experiments with the Unitree A1 quadruped robot, we demonstrate that a single RL policy yielded by \our can generalize to various velocity and posture command combinations on both structured and unstructured terrains, achieving agile posture-aware locomotion.

\begin{figure}[tb]
    \centering
    \vspace{5pt}
    \includegraphics[width=\linewidth]{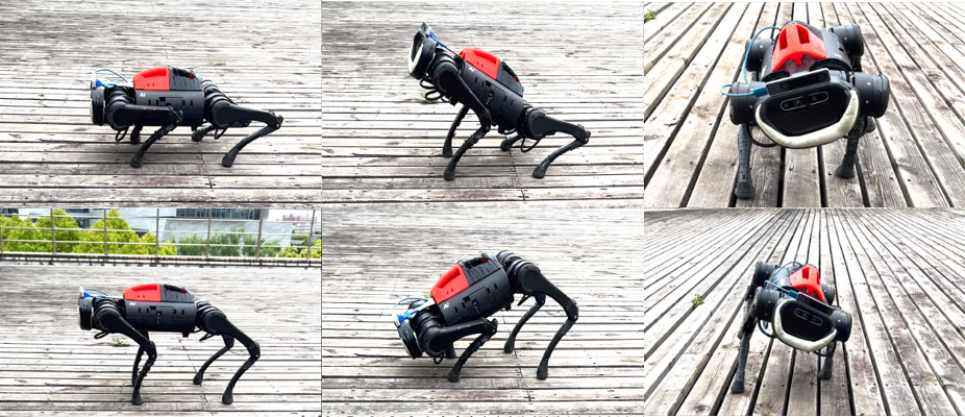}
    \vspace{-15pt}
    \caption{Unitree A1 quadruped robot demonstrating posture-aware locomotion control (from left to right: height, pitch, and roll control)}
    \vspace{-15pt}
    \label{fig:demo}
\end{figure}

The key contributions of this research can be summarized as follows:

\begin{enumerate}
    \item We propose \our, an end-to-end deep reinforcement learning framework that integrates body velocity and posture control, thus significantly enhancing locomotion controllability and adaptability.
    \item We introduce a customized training curriculum, which combines terrain, reward, and command curriculum learning. This enables quadruped robots to follow a 6-dimensional (6D) command for posture-aware locomotion control.
    \item Our real-world experiments demonstrate that \our successfully generalizes to various environments without additional fine-tuning, effectively bridging the sim-to-real gap and enabling robust deployment.
\end{enumerate}

\section{RELATED WORK}

\subsection{Quadrupedal Locomotion with Reinforcement Learning}

In recent years, deep reinforcement learning has become increasingly popular for quadrupedal locomotion control due to its ability to learn complex behaviors in dynamic environments. One of the most common focuses in this domain is velocity control. By directly controlling the robot’s linear and angular velocity, these methods enable the robot to navigate towards desired destinations \cite{smith_walk_2022, ji_concurrent_2022, fu_coupling_2022, nahrendra_dreamwaq_2023, gangapurwala_guided_2020, hwangbo_learning_2019, miki_learning_2022, agarwal_legged_2022, kumar_rma_2021, lai_sim--real_2023}. However, pure velocity control often lacks flexibility, especially when navigating complex terrains or performing tasks that require more nuanced body movements. To address this limitation, some studies have expanded the system to include gait control \cite{bellegarda_cpg-rl_2022, yang_multi-expert_2020, bellegarda_visual_2024, margolis_walk_2023}, where the focus is on adjusting the robot’s leg movements to generate desired motion patterns, such as walking, trotting, or running.

Despite the progress in gait-adaptive locomotion, there are still challenges. One major issue is the coordination of posture and movement, especially in more dynamic or unpredictable environments. While gait control helps with the rhythm of movement, it does not directly address the robot’s posture — specifically the orientation and height of its body during movement. For example, the robot may need to adjust its posture to maintain body stability when traversing slopes or stairs, but traditional gait control does not inherently solve these problems.

This is where explicit posture tracking becomes critical. While few studies have explicitly focused on posture tracking in quadruped robots, our approach integrates posture-aware locomotion control using reinforcement learning and customized curricula, allowing the robot to simultaneously adapt its posture and movements to explore complex environments.

\subsection{Adversarial Motion Prior}

Designing effective reward functions for reinforcement learning (RL) remains a significant challenge, particularly in domains such as robot locomotion where intricate physical dynamics must be navigated. Crafting rewards that balance task performance with motion quality often demands substantial domain expertise and iterative tuning. Overly complex reward structures, while aiming to encode desired behaviors, can inadvertently lead to unstable or unnatural movement patterns, as agents may exploit reward loopholes through unintended strategies. To address this, researchers have introduced regularization techniques \cite{kumar_rma_2021, margolis_rapid_2022, agarwal_legged_2022, yang_learning_2021, ji_concurrent_2022} to constrain policy optimization, promoting structured and energy-efficient gaits. These methods often incorporate penalties for deviations from biomechanically plausible joint motions or incentives for symmetry and periodicity, thereby reducing the need for exhaustive reward engineering.



Recent advances have explored adversarial motion priors (AMP) as a means to implicitly guide policies toward naturalistic behaviors. By leveraging generative adversarial imitation learning (GAIL) frameworks \cite{ho_generative_nodate}, studies such as those by Escontrela et al. \cite{escontrela_adversarial_2022} and Wu et al. \cite{wu_learning_2023, wu_learning_2023-1} have demonstrated the efficacy of adversarial training for motion synthesis. In this paradigm, a discriminator network is trained to distinguish between agent-generated state transitions and those from an expert dataset, producing a style reward that incentivizes policy behaviors to align with the expert distribution. This approach circumvents the need for manually specified reward terms by instead deriving supervision directly from expert demonstrations, enabling agents to learn coherent, human-like locomotion without explicit kinematic constraints. 


\subsection{Sim-to-Real Transfer}

Deploying policies trained in simulation to real-world robotic systems continues to face substantial hurdles due to the sim-to-real gap \cite{lai_sim--real_2023, zhao_sim--real_2020, tan_sim--real_2018}. Discrepancies in dynamic modeling, imperfect terrain representations, unaccounted physical interactions (e.g., friction, material deformation), and mismatches in sensor noise or actuation delays often degrade policy performance when transferring to hardware. These challenges necessitate strategies to bridge the divide between simulated training environments and real-world operational conditions.


A common methodology involves high-fidelity simulation \cite{tan_sim--real_2018}, which meticulously replicates real-world physics, sensor behaviors, and environmental textures—including terrain variability, actuator dynamics, and communication delays—to reduce modeling inaccuracies. Alternatively, domain randomization \cite{escontrela_adversarial_2022, ji_concurrent_2022, wu_learning_2023, yang_learning_2021, kim_not_2023, margolis_rapid_2022, gangapurwala_rloc_2022, kumar_rma_2021, zhuang_robot_2023, tan_sim--real_2018, margolis_walk_2023} deliberately introduces variability during training by perturbing parameters such as robot inertia, joint damping, ground friction, and sensor noise ranges. By exposing policies to diverse simulated conditions, this approach fosters robustness, enabling adaptation to unseen real-world dynamics without overfitting to idealized simulation properties. Together, these strategies aim to narrow the sim-to-real gap, though balancing realism and computational efficiency remains an active area of research.


\section{Methodology}

\subsection{Reinforcement Learning Formulation}
RL addresses the problem of sequential decision-making. Given that the robot only receives partial environmental information filtered by proprioceptive sensors, we model locomotion control as a Partially Observable Markov Decision Process (POMDP), defined by the tuple ($\mathcal{S}, \mathcal{O}, \mathcal{A}, \mathcal{P}, r, \gamma$), where \(\mathcal{S}\), \(\mathcal{O}\), and \(\mathcal{A}\) are the state, observation, and action spaces, respectively. \(\mathcal{P}: \mathcal{S} \times \mathcal{A} \times \mathcal{S} \rightarrow \mathbb{R}\) is the state transition probability, \(r: \mathcal{S} \times \mathcal{A} \rightarrow \mathbb{R}\) is the reward function, and \(\gamma \in [0, 1)\) is the discount factor.

An episode begins from an initial state \(s_0\). At each time step \(t\), the state \(s_t\) contains a complete description of the environment. During simulator training, the command signal \(c_t\) is given randomly, whereas during real-world deployment, it is issued by a human operator via remote control. The agent takes an action \(a_t = \pi(o_t, c_t)\) based on the current observation \(o_t\) and command signal \(c_t\), and interacts with the environment. The environment transitions to the next state \(s_{t+1}\) with probability \(\mathcal{P}(s_{t+1} \mid s_t, a_t)\), providing the agent with a reward \(r_t\). The objective of RL is to find the optimal policy \(\pi_\theta\) that maximizes the expected return:
\begin{equation}
\underset{\theta}{\text{maximize}} \; \mathbb{E}_{\pi_{\theta}} \left[ \sum_{t=0}^{\infty} \gamma^t r_t \right].
\end{equation}

\subsubsection{Observation Space}

\begin{figure}[tb]
\centering
\vspace{3pt}
\includegraphics[width=0.48\textwidth]{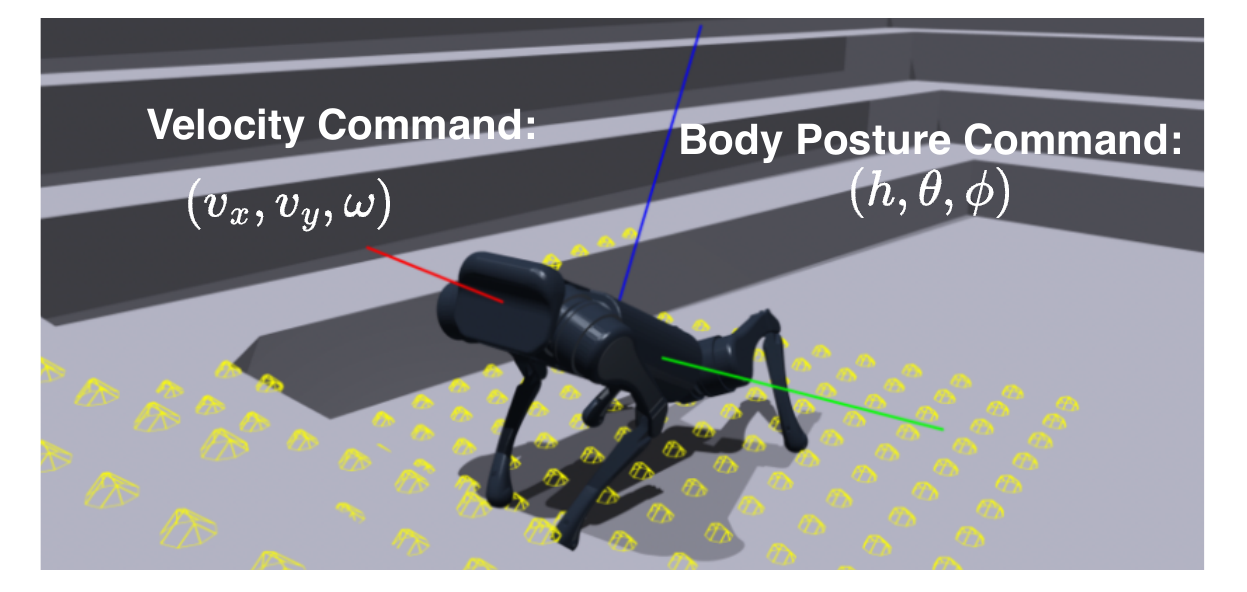}
\vspace{-15pt}
\caption{Schematic diagram of the robot's body frame rendered in simulation, showing the coordinate axes (red: x-axis, green: y-axis, blue: z-axis) and local height map sampling points (yellow). The 6D commands correspond to the robot's motion tracking in linear velocity, angular velocity, height, pitch, and roll.}
\vspace{-10pt}
\label{fig:robot_simulation}
\end{figure}

Proprioceptive observation \(o_t \in \mathbb{R}^{50}\) includes angular velocity around the z-axis in the robot base coordinate system, body pitch and roll angles, gravity projection, 6D commands, DOF positions and linear velocities in the current time step, and DOF positions in the previous time step. The 6D commands consist of linear velocity in the x and y directions, angular velocity around the z-axis, body height, pitch, and roll tracking (see Fig.~\ref{fig:robot_simulation}). The robot's posture angles are derived from the Inertial Measurement Unit (IMU) measurements, with the quaternion converted to Euler angles for alignment with commands. 

Assuming the robot's quaternion is represented as \(q = w + xi + yj + zk\), where \(w\) is the real part, and \((x, y, z)\) are the imaginary parts, the corresponding Euler angles are:
\begin{equation}
\begin{aligned}
\phi &= \text{atan2}(2(w x + y z), 1 - 2(x^2 + y^2)), \\
\theta &= \text{asin}(2(w y - z x)), \\
\psi &= \text{atan2}(2(w z + x y), 1 - 2(y^2 + z^2)),
\end{aligned}
\end{equation}
where \(\phi\) represents the roll angle, \(\theta\) represents the pitch angle, and \(\psi\) represents the yaw angle.

\subsubsection{Action Space}

The action \(a_t \in \mathbb{R}^{12}\) specifies the target position for each DOF, converted to joint torques via a Proportional-Derivative (PD) controller:
\begin{equation}
\tau = k_p (q_d - q) + k_d (\dot{q}_d - \dot{q}),
\end{equation}
where \(q\) and \(q_d\) are the current and target DOF positions and \((k_p, k_d)\) are the hyperparameters of the PD controller.

\subsubsection{Reward Function}

Our goal is to train an agile and robust blind locomotion controller. The reward function includes terms to encourage locomotion tasks and prevent undesirable behaviors. Inspired by [14, 16, 34], we use the AMP to replace complex auxiliary reward terms with style rewards, aiming for a simple and effective reward function:
\begin{equation}
r_t = r_t^{task} + r_t^{style} + r_t^{reg}.
\end{equation}

The task reward encourages the robot to follow the 6D commands, including linear and angular velocities, body height, pitch, and roll. The reward for linear velocity tracking is defined as:
\begin{equation}
R_{v} = \exp\left(-\frac{\|v_{actual} - v_{cmd}\|^2}{\sigma_v}\right),
\end{equation}
where \(v_{actual}\) and \(v_{cmd}\) are the actual and commanded linear velocities. \(\sigma_v\) is a scaling factor reflecting the reward's sensitivity to velocity error. The smaller \(\sigma_v\), the more sensitive the reward is to the velocity error. Similarly, the reward for angular velocity tracking is as follows:
\begin{equation}
R_{\omega} = \exp\left(-\frac{\|\omega_{actual} - \omega_{cmd}\|^2}{\sigma_\omega}\right).
\end{equation}

Additional reward terms control body posture (height and orientation):
\begin{align}
R_{h} &= \exp\left(-\frac{|h_{actual} - h_{cmd}|}{\sigma_h}\right), \\
R_{\theta} &= \exp\left(-\frac{\|\theta_{actual} - \theta_{cmd}\|^2}{\sigma_\theta}\right).
\end{align}

The style reward encourages imitation of expert behaviors, aiding in transferring stable gaits to complex terrains. Following  \cite{escontrela_adversarial_2022}, the AMP style reward is defined as:
\begin{equation}
r_t^{style}(s_t,s_{t+1}) = \text{max}[0, 1-0.25(D_{\varphi}(s,s')-1)^2],
\end{equation}
where \(D_{\varphi}(s, s') \in [-1, 1]\) is the discriminator output comparing agent behavior to expert data. Other regularization terms are similar to  \cite{wu_learning_2023}. 



\subsection{AMP Motion Data Generation}

There are two modes for generating reference expert motion datasets. The first involves using motion capture technology to collect animal motion data, then training robots to mimic these behaviors \cite{escontrela_adversarial_2022}. The second collects motion data directly from quadruped robots. \our adopts the latter approach. Initially, the robot is trained on flat terrain without involving AMP, using auxiliary reward terms to regularize motion. Once this training stage is complete, AMP data is collected from the robot's learned policy interactions. This stable gait data from flat terrain can be reused for training on more complex terrains. In the next stage, AMP is combined to train the robot in diverse, complex terrains, enabling stable traversal even on challenging surfaces.

\subsection{Asymmetric Actor-Critic}

 Privileged learning \cite{chen_learning_2020} improves policy adaptability by leveraging simulation-only information during training. Recent works leverage a two-stage Teacher-Student paradigm \cite{lee_learning_2020, wu_learning_2023, miki_learning_2022, agarwal_legged_2022, kim_not_2023, margolis_rapid_2022, kumar_rma_2021, lai_sim--real_2023}, where a teacher policy is trained in simulation using privileged information, and a student policy is distilled for real-world deployment. However, this approach suffers from low data efficiency, and the student policy often underperforms the teacher \cite{nahrendra_dreamwaq_2023}.

\begin{figure}[tb]
\centering
\vspace{5pt}
\includegraphics[width=1\linewidth]{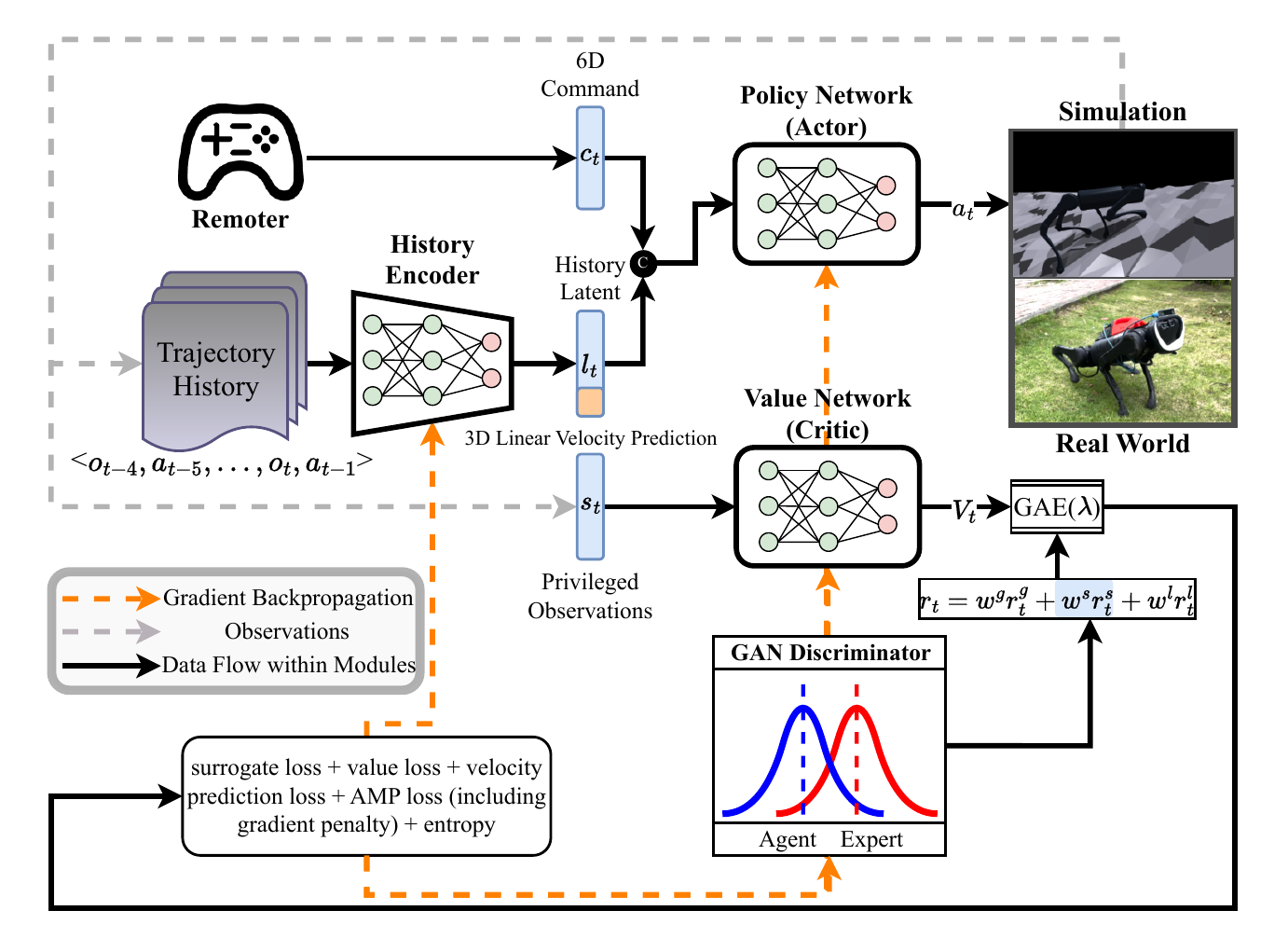}
\vspace{-15pt}
\caption{Overview of our framework \our. The orange dashed line represents gradient backpropagation during training, while the gray dashed line indicates observations obtained from the simulator or real-world environment and fed into our framework. Solid arrows denote the data flow within our model. The actor is a shallow MLP that outputs 12D target DOF positions, which are converted to torques via a PD controller. The actor’s input combines a 6D command with a history latent vector from a history encoder. The critic is a shallow MLP, and the AMP discriminator is an MLP.}
\vspace{-5pt}
\label{fig:method_overview}
\end{figure}

Actor-critic algorithms improve efficiency by combining value function approximation with policy gradient. The actor (policy network) interacts with the environment, while the critic (value network) evaluates state values and guides policy updates. To enhance data efficiency under partial observability, we adopt an asymmetric actor-critic architecture \cite{pinto_asymmetric_2017, yue_aacc_2022, zhang_resilient_2024, nahrendra_dreamwaq_2023, wu_learning_2023-1}, as illustrated in Fig.~\ref{fig:method_overview}. The critic receives full observations, including privileged information, while the actor relies solely on proprioceptive input. This enables the critic to provide richer learning signals, improving policy performance.

\our encodes the last five timesteps of observations and actions into a latent vector via a historical encoder. This vector, concatenated with 6D commands, serves as the actor's input to generate 12D target joint positions. These positions are converted into motor torques by a PD controller. Meanwhile, the critic receives privileged inputs, including ground heights, friction, restitution coefficient, base mass, center of mass (CoM) position, contact forces, and Boolean contact indicators.

At each time step \(t\), the actor performs an action based on partial observations, and the environment returns new observations and rewards. The critic evaluates the state using privileged information and provides value estimates for policy updates. This process enables the actor to implicitly leverage privileged information during training, improving robustness under real-world constraints.

\subsection{Training Loss}

We use the Proximal Policy Optimization (PPO) algorithm with clipping range \cite{schulman_proximal_2017}. Its objective is:


\begin{align}
L^{CLIP}&(\theta') = \mathbb{E}_{s, a} \Bigg[  \min \Bigg(\frac{\pi_{\theta'} \left( a \mid s \right)}{\pi_{\theta} \left( a \mid s \right)} A^{\pi_\theta} \left( s, a \right), \nonumber\\
& \text{clip} \left( \frac{\pi_{\theta'} \left( a \mid s \right)}{\pi_{\theta} \left( a \mid s \right)}, 1 - \epsilon, 1 + \epsilon \right) A^{\pi_\theta} \left( s, a \right) \Bigg) \Bigg],
\end{align}

where \( A^{\pi_\theta} \left( s, a \right) \) is the advantage function, which measures the return of taking action \( a \) in state \( s \) compared to the average expected return. This approach reduces the risk of "policy collapse" by limiting policy updates, and preventing significant performance drops due to large policy changes.

The critic network's loss function minimizes the Mean Square Error (MSE) between estimated and actual returns:
\begin{equation}
L^{VF}(\phi) = \frac{1}{2} \mathbb{E}_{\left( s, a \right) \sim \pi_{\theta}} \left[ (V_\phi(s) - G_t)^2 \right].
\end{equation}

For the AMP discriminator, we adopt the training objective from \cite{escontrela_adversarial_2022}, minimizing the AMP loss function:
\begin{equation}
\begin{aligned}
L^{AMP}(\varphi) &= \mathbb{E}_{\left( s, s' \right) \sim \mathcal{D}} \left[ \left( D_{\varphi}(s, s') - 1 \right)^2 \right]\\ 
&+ \mathbb{E}_{\left( s, s' \right) \sim \pi_{\theta} \left( s, a \right)} \left[ \left( D_{\varphi}(s, s') + 1 \right)^2 \right]\\
&+ \frac{\omega^{gp}}{2} \mathbb{E}_{\left( s, s' \right) \sim \mathcal{D}} \left[ \left\| \nabla_{\varphi} D_{\varphi}(s, s') \right\|^2 \right],
\end{aligned}
\end{equation}
where the first two terms help distinguish state transitions from the reference expert dataset and the agent's policy. The last term is a gradient penalty that prevents overfitting and improves the stability of the training \cite{escontrela_adversarial_2022}.

\subsection{Training Curricula}

To stabilize early RL, we employ curriculum learning \cite{lesort_continual_2019} to gradually increase task difficulty based on policy performance. Our curricula include terrain, reward, and command curricula.

\subsubsection{Terrain Curriculum}

We construct diverse simulated terrains --- wavy surfaces, rough slopes, ascending and descending stairs, discrete obstacles, and flat rough terrain \cite{rudin_learning_2022} --- to mimic urban and wilderness environments. Initially, all agents start on the easiest terrains. Successful agents progress to more difficult terrains, while those struggling are reset to easier terrains. Adaptability is measured by the distance traveled. Agents traveling more than half the terrain size move to more difficult terrains, while those traveling less are reset to easier terrains.

\subsubsection{Reward Curriculum}

We design a reward curriculum for posture control to gradually increase task complexity. Early training prioritizes velocity tracking, mid-training gradually introduces body height, pitch, and roll tracking, and later training applies the full reward function for integrated locomotion and posture control.

\subsubsection{Command Curriculum}

Commands are resampled periodically for diverse scenarios. Except for ascending and descending stairs, pitch and roll commands follow a normal distribution \(\mathcal{N}(0, 0.25)\), while body height commands are uniformly sampled from 0.1 meters to 0.4 meters. Pitch angles are constrained to \([-\frac{\pi}{4}, \frac{\pi}{4}]\) radians and roll angles to \([-\frac{\pi}{6}, \frac{\pi}{6}]\) radians.

For stair traversal, orientation angle distributions are:
\begin{itemize}
    \item \textbf{Ascending stairs}: Pitch commands follow a normal distribution with a mean of -0.5 radians and a standard deviation of 0.25 radians, clipped to \([-\frac{\pi}{4}, 0]\) radians.
    \item \textbf{Descending stairs}: Pitch commands follow a normal distribution with a mean of 0.5 radians and a standard deviation of 0.25 radians, clipped to \([0, \frac{\pi}{4}]\) radians.
\end{itemize}

Locomotion under 6D command constraints is challenging due to physical limitations. We employ a grid adaptive curriculum \cite{margolis_rapid_2022}, dynamically expanding velocity ranges and reducing disturbance intervals when the velocity tracking reward exceeds 80\% of its maximum.

\subsection{Domain Randomization}

To bridge the sim-to-real gap, we apply domain randomization \cite{tobin_domain_2017}, varying training conditions to cover real-world uncertainties. Domain randomization consists of visual and dynamics randomization \cite{zhao_sim--real_2020, xie_dynamics_2021}. Since this paper focuses on low-level blind locomotion control in quadruped robots, we apply dynamics randomization, adjusting robot dynamics, environment parameters, and sensor noise.

We randomize ground friction, restitution, base and link mass, action delay, PD controller gains, and motor torque. Additionally, we introduce random velocity disturbances to simulate external force perturbations, enhancing robustness. The randomized variables and their sampling ranges are detailed in Table~\ref{tab:randomized_variables}.

\begin{table}[tb]
\centering
\vspace{10pt}
\caption{Randomized Variables and Sampling Ranges}
\begin{tabular}{|c||c||c||c|}
\hline
Variable & Minimum & Maximum & Unit \\
\hline
Ground friction & 0.05 & 2.75 & - \\
Restitution & 0 & 1.0 & - \\
Load mass & 0 & 3.0 & kg \\
Link mass & 80 & 120 & \% \\
CoM position & -0.05 & 0.05 & m \\
PD controller P-gain & 80 & 120 & \% \\
PD controller D-gain & 80 & 120 & \% \\
Motor power & 80 & 120 & \% \\
Action delay & 0.00 & 0.02 & s \\
\hline
\end{tabular}
\vspace{-5pt}
\label{tab:randomized_variables}
\end{table}

\section{Results and Discussion}

In this section, we present the results of our experimental setup and provide a detailed analysis. Our experiments aim to address the following key questions:
\begin{itemize}
    \item Can \our enable the robot to effectively track 6D control commands to manage locomotion and posture?
    \item How well does the policy trained in the simulator transfer to real-world environments with complex terrains?
    \item Which components are critical for \our's overall performance?
\end{itemize}

To explore these questions, we first trained the robot's locomotion controller in the simulator and then evaluated its command tracking performance. Following this, the policy was transferred to a real robot for deployment in real-world conditions. A series of ablation studies were also conducted to identify the most significant factors affecting the performance of the algorithm.

Regarding hardware details, all real-world experiments were conducted using the Unitree A1 quadruped robot \cite{noauthor_unitree_nodate}. This quadruped robot, weighing approximately 12 kilograms, features 18 DOF, 12 of which are actuated (three motors per leg).

\subsection{Simulation Experiments}
\label{sec:experiments_sim}

We used the Isaac Gym simulator \cite{noauthor_210810470_nodate} for both training and experimentation, employing the Proximal Policy Optimization (PPO) \cite{schulman_proximal_2017} algorithm with 4096 agents trained in parallel across different terrain types. The training process, conducted on a single NVIDIA GeForce RTX 4090 GPU, took 32.8 hours, equivalent to 284 days of real-world training. The Adam optimizer was used, and the control frequency in the simulator was set to 200 hertz, with each reinforcement learning episode running for up to 4000 timesteps (20 seconds in real-time). Early episode termination would be triggered by collisions.

The robot was tested on various terrains to assess its 6D command tracking capability. The commands received by the robot were represented as a six-tuple: linear velocity along the x-axis (\(v_x\)), linear velocity along the y-axis (\(v_y\)), angular velocity around the z-axis (\(\omega_z\)), body height (\(h\)) (the desired height difference from the reference height), pitch angle (\(\theta\)), and roll angle (\(\phi\)). The units for linear velocity, angular velocity, height and angles are meters per second, radians per second, meters, and radians, respectively.

\begin{figure}[tb]
\centering
\vspace{5pt}
\includegraphics[width=0.48\textwidth]{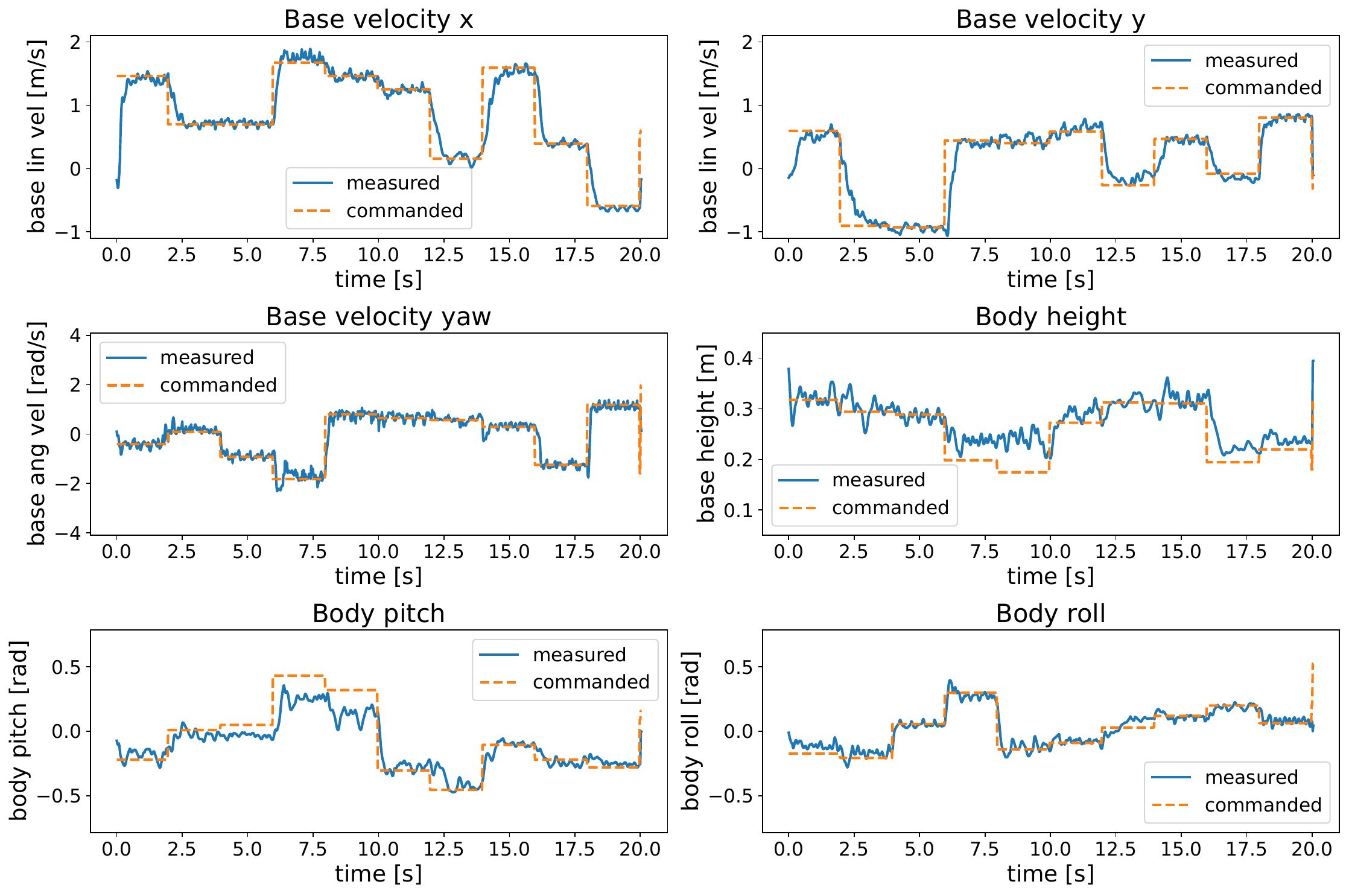}
\vspace{-15pt}
\caption{Dynamic 6D command tracking performance on simulated flat terrain.}
\vspace{-5pt}
\label{fig:test_sim_flat}
\end{figure}

In order to demonstrate the robot’s command tracking capability under dynamic conditions, we performed testing primarily on flat terrain, where 6D control commands were dynamically resampled every 2 seconds. The results (shown in Fig.~\ref{fig:test_sim_flat}) indicate that the robot can effectively track a variety of control commands, adjusting its body posture in real-time whether stationary or moving. This includes combined actions such as walking, crawling, in-place rotation, and posture adjustments.

As observed in the figure, the robot shows excellent tracking performance for \(v_x\), \(v_y\), yaw rate, pitch, and roll, with very minimal tracking error throughout the entire test. This highlights the robot’s robust ability to adapt to a range of control commands and execute them effectively. However, we also noticed that the height command tracking showed some degradation during the interval between 6-10 seconds. During this period, the robot was commanded with unconventional (non-neutral) posture angles that conflicted with the height commands, which requested the robot to crawl. This discrepancy in the height response illustrates the robot’s adaptability, as it handled the conflicting command combinations effectively. It also reflects the flexibility and robustness of the reinforcement learning policy, which was able to maintain stable performance even in the presence of command conflicts.




While the focus of our analysis is on flat terrain, the robot was also tested on more complex terrains, such as wavy surfaces, slopes, terrain with discrete obstacles, and stairs. The robot demonstrated consistent and robust tracking performance across all environments. On wavy terrain, the robot accurately tracked velocity and posture angle commands, with only slight instability observed in height tracking due to surface fluctuations. On sloped surfaces, the robot maintained a stable body posture, and on stairs, it successfully adapted to height changes, maintaining its posture while climbing and descending.

The primary reason for focusing on flat terrain in this section is its ability to fully capture the robot’s performance in dynamic command tracking, providing a comprehensive illustration of the system’s capabilities.

\subsection{Real-World Experiments}



After training the policy in simulation, we directly deployed it onto the real A1 robot without any additional fine-tuning. The control frequency was set to 50 Hz, with the PD controller parameters configured as $k_p=28$ and $k_d=0.7$.

\begin{figure}[tb]
\centering
\vspace{10pt}
\includegraphics[width=0.48\textwidth]{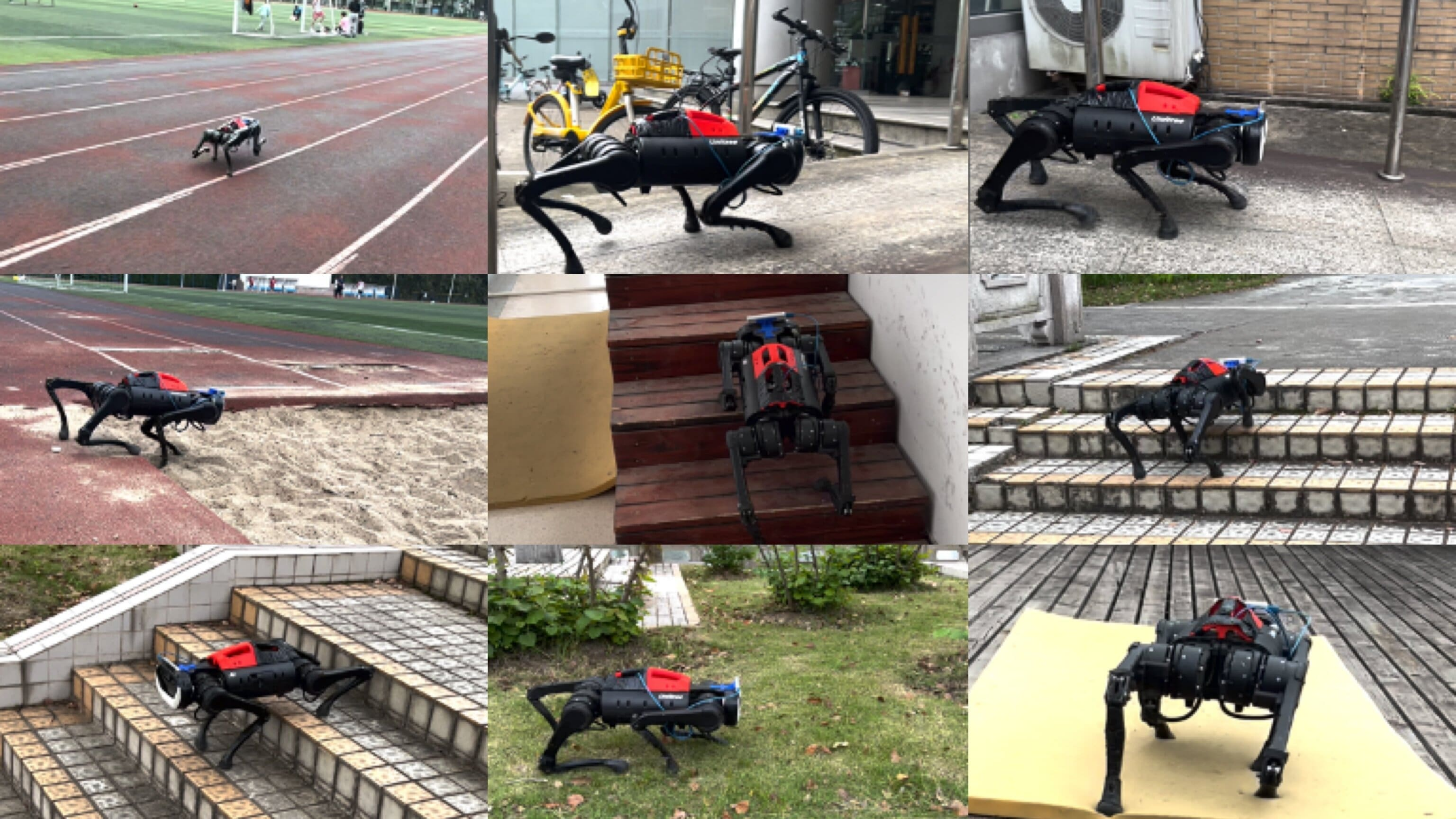}
\vspace{-15pt}
\caption{Real-world deployment of the trained policy on various structured and unstructured terrains. The 3×3 composite image illustrates the robot navigating different terrain types, demonstrating its adaptability.}
\label{fig:real-world_traversal}
\end{figure}

To evaluate the robot’s locomotion capabilities, we conducted tests on nine different terrains: grass, rubber tracks, soft sponge pads, uphill slopes, downhill slopes, sand, wooden stairs (ascending), wooden stairs (descending), and step traversal (both up and down). Each test was repeated five times for consistency. The results demonstrated that the robot successfully navigated all terrains, achieving a 100\% success rate except for wooden stair ascent, where only one failure was observed due to a collision with a step, which triggered our predefined safety protocol. A visual overview of the real-world experiments is presented in Fig.~\ref{fig:real-world_traversal}, which showcases the robot traversing these diverse terrains. Additional demonstrations of posture-aware locomotion tracking can be found in the accompanying video.

\begin{figure}[tb]
\centering
\includegraphics[width=0.48\textwidth]{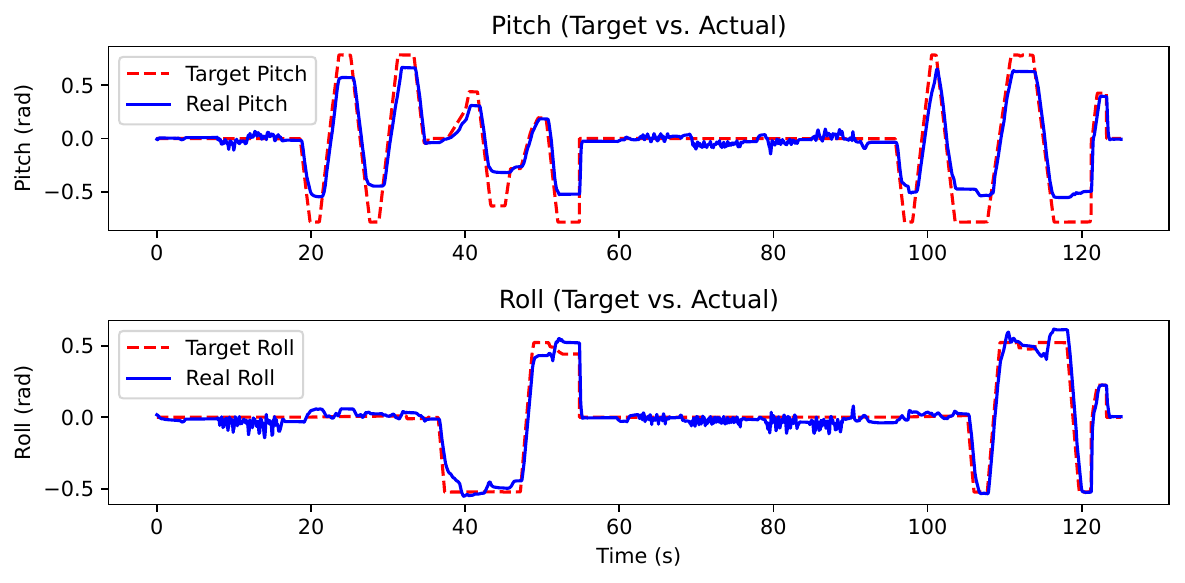}
\vspace{-15pt}
\caption{Real-time tracking of pitch and roll angles. The figure compares target (red dashed line) and actual (blue solid line) angles, demonstrating the robot's posture control capabilities during real-world operation.}
\vspace{-5pt}
\label{fig:test_real}
\end{figure}

In addition to evaluating the robot’s locomotion across various terrains, we also collected data on the real-time tracking of pitch and roll angle commands. Here we focus on demonstrating the robot’s body orientation control capabilities, as body height in the real world is difficult to measure due to the absence of specific sensors. The angle measurements are derived from IMU data and converted to Euler angles. The tracking results, shown in Fig.~\ref{fig:test_real}, highlight the robot’s ability to track target pitch and roll angles in real time. From the figure, we can observe that \our successfully overcomes the sim-to-real challenge, enabling fast and stable responses to posture commands. The robot shows excellent tracking performance, adapting well to real-world conditions.

\subsection{Ablation Study}

Through ablation experiments in the simulation, we assessed the contributions of different components to \our's performance. Key findings include the significant impact of AMP technology, our tailored training curricula and our sampling strategy for body posture commands on overall performance.

\subsubsection{Effect of AMP}
\label{sec:experiments_amp}





We compared the performance of policies trained with and without AMP, focusing on average task rewards and body collision counts across flat and complex terrains.

\begin{figure}[tb]
\centering
\vspace{5pt}
\includegraphics[width=0.48\textwidth]{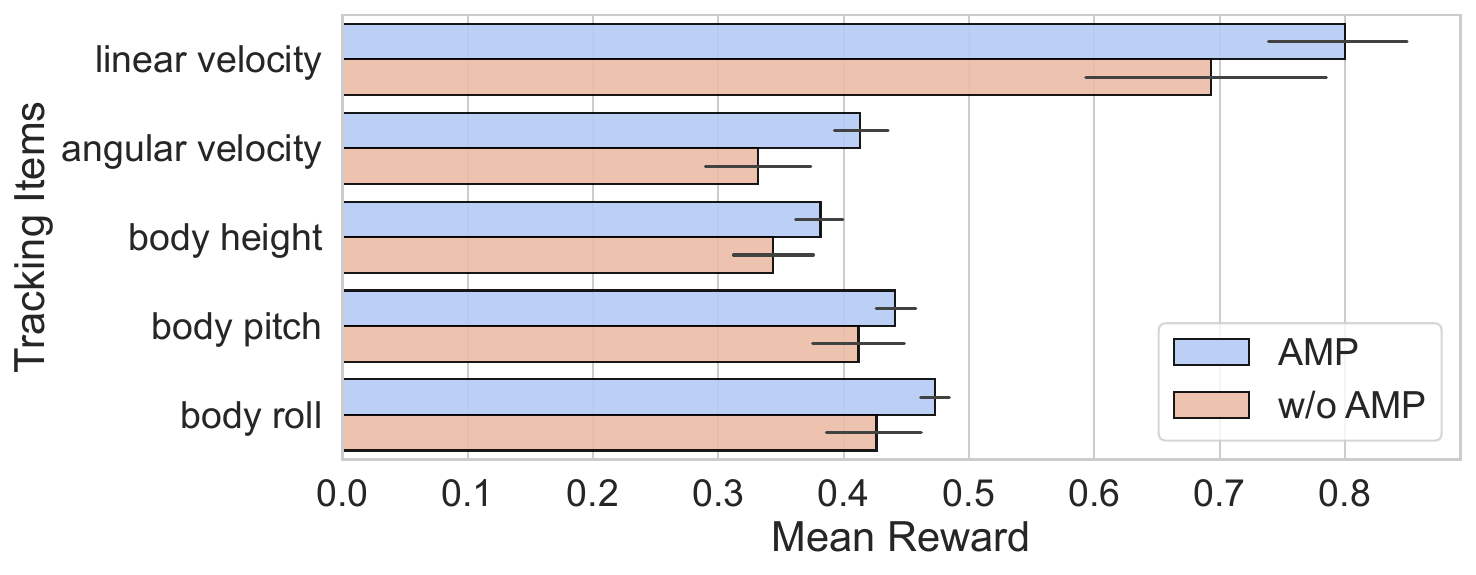}
\vspace{-15pt}
\caption{Performance comparison between policies trained with and without AMP.}
\vspace{-5pt}
\label{fig:ablation_amp}
\end{figure}

For flat terrain, dynamic commands were used, as described in Sec.\ref{sec:experiments_sim}. On more complex terrains, we employed static commands (1.0, 0.0, 0.0, 0.0, 0.0, 0.0) to isolate the effect of AMP. The results were derived from five tests using the same random seed, with mean and standard deviation calculated for each comparison. The average reward data is presented in Fig.\ref{fig:ablation_amp}, illustrating AMP’s impact on policy performance across terrain types.

While AMP had a minimal effect on flat terrain, it substantially improved performance on complex terrains. Specifically, on terrains with discrete obstacles and stairs, the robot without AMP experienced multiple collisions, often resulting in episode terminations and near-zero task completion rates. However, with AMP, the robot successfully completed all tests without body collisions causing any terminations.

\subsubsection{Customized Training Curricula}


Curriculum learning has been widely used to improve RL policy performance \cite{nahrendra_dreamwaq_2023, lee_learning_2020, wu_learning_2023, miki_learning_2022, kim_not_2023, margolis_rapid_2022, zhuang_robot_2023}. \our innovates by introducing a staged reward curriculum to improve the robot’s ability to navigate complex terrains while controlling its posture. In particular, we evaluate the impact of the reward curriculum on policy performance.

The results show that the reward curriculum has a significant impact on performance across various terrains. For example, when testing on uphill stair terrains, the reward curriculum reduced collision rates and improved mobility. This finding confirms that a carefully designed reward function is critical for optimizing RL policy performance. Specifically, periodically adjusting the composition and magnitude of reward signals during training improves the robot’s ability to handle multiple task objectives.


\subsubsection{Sampling Strategy}

We also examined the effects of different command sampling strategies during simulator training. These included:
\begin{itemize}
    \item Sampling posture control commands from a normal distribution, and sampling velocity commands from a uniform distribution.
    \item Sampling both posture and velocity commands from a uniform distribution.
\end{itemize}

\begin{figure}[tb]
\centering
\vspace{5pt}
\includegraphics[width=0.48\textwidth]{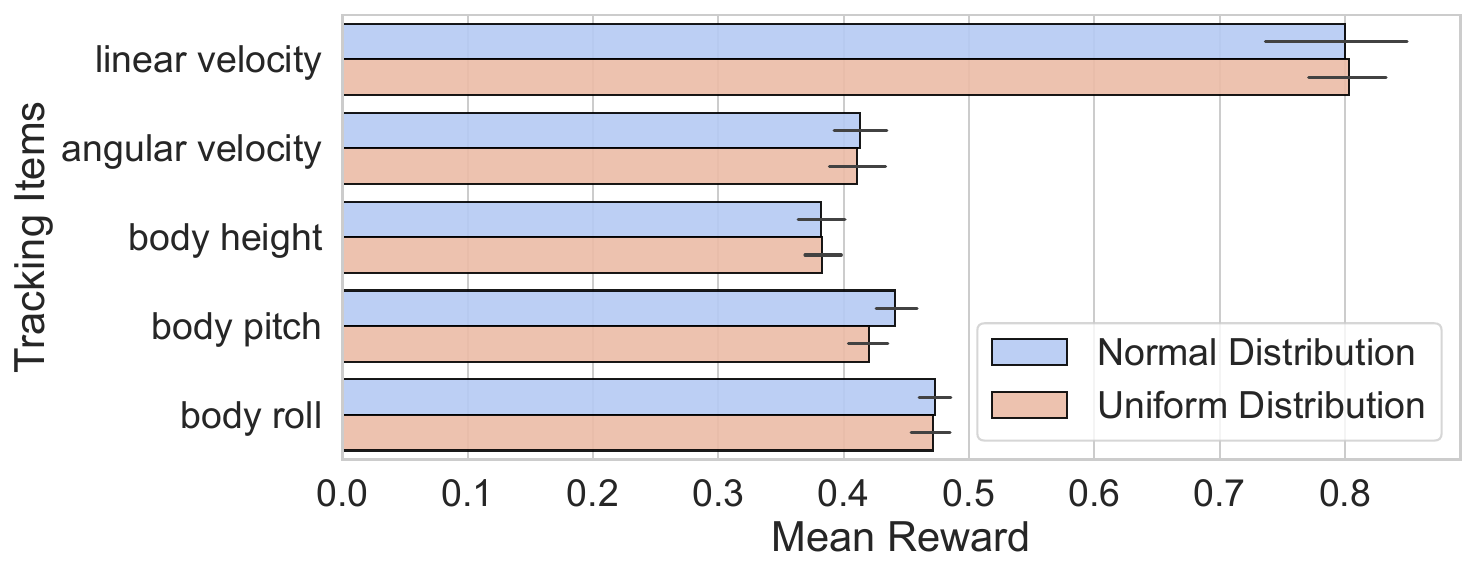}
\vspace{-15pt}
\caption{Performance comparison between policies trained with normal distribution sampling and uniform distribution sampling for posture commands.}
\label{fig:ablation_sampling}
\end{figure}

The results (Fig.~\ref{fig:ablation_sampling}) indicated that normally distributed sampling for angular velocity and body posture commands led to slightly better performance in orientation tracking compared to uniformly distributed sampling, although the overall differences in average reward were marginal. However, it is worth noting that with the uniform distribution sampling, two episodes were prematurely terminated due to collisions during testing.

\subsubsection{Historical Encoder Architectures}

We compared three different historical encoder architectures: 
\begin{itemize}
    \item MLP-based encoder using the last 5 timesteps of observations (MLP-5).
    \item MLP-based encoder using the last 50 timesteps of observations (MLP-50).
    \item 1D-Temporal Convolutional Network using the last 50 timesteps (TCN-50).
\end{itemize}

\begin{figure}[tb]
\centering
\includegraphics[width=0.48\textwidth]{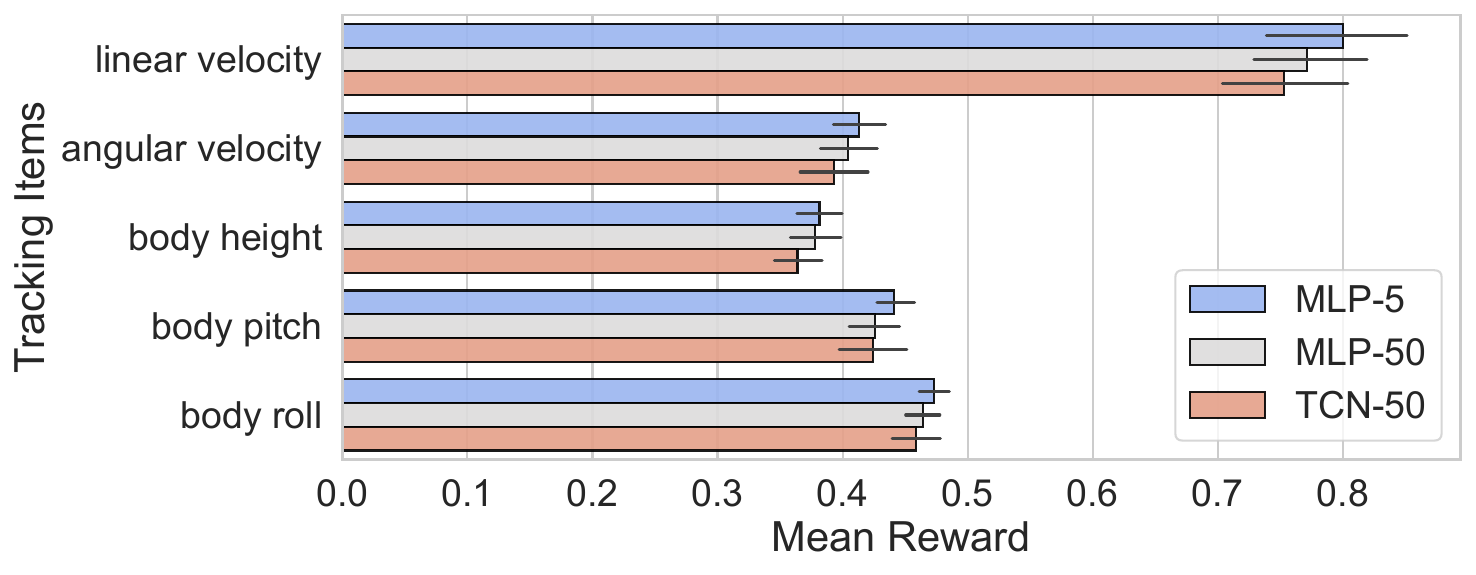}
\vspace{-15pt}
\caption{Performance comparison of different historical encoder architectures.}
\vspace{-5pt}
\label{fig:ablation_encoder_architecture}
\end{figure}

The results (Fig.~\ref{fig:ablation_encoder_architecture}) revealed that the MLP-5 architecture, with its shorter history of observations, slightly outperformed the others. This suggests that short-term memory is sufficient for controlling low-level blind locomotion, and a simpler network structure may be more effective for this task.

\section{Conclusions}



In this work, we present a deep reinforcement learning framework for posture-aware locomotion control of quadruped robots named \our. Our approach focuses on achieving robust and simultaneous locomotion and posture control across a wide range of terrains, without relying on visual information. This enables quadruped robots to acquire a variety of motion skills, significantly enhancing their adaptability and intelligence. By leveraging \our, we improve the robot’s ability to effectively navigate complex and dynamic environments.

Looking ahead, we aim to extend this work by developing upper-level modules, such as motion planning systems, to facilitate intelligent navigation, obstacle avoidance, and high-level decision-making. Additionally, we plan to integrate large language models into future iterations of the system, enabling more natural and sophisticated human-robot interaction. This will allow robots to interpret and respond to high-dimensional commands, further advancing their autonomy and practical deployment in real-world applications.









\bibliographystyle{IEEEtran}  
\bibliography{references}  

\begin{thebibliography}{10}
\providecommand{\url}[1]{#1}
\csname url@samestyle\endcsname
\providecommand{\newblock}{\relax}
\providecommand{\bibinfo}[2]{#2}
\providecommand{\BIBentrySTDinterwordspacing}{\spaceskip=0pt\relax}
\providecommand{\BIBentryALTinterwordstretchfactor}{4}
\providecommand{\BIBentryALTinterwordspacing}{\spaceskip=\fontdimen2\font plus
\BIBentryALTinterwordstretchfactor\fontdimen3\font minus \fontdimen4\font\relax}
\providecommand{\BIBforeignlanguage}[2]{{%
\expandafter\ifx\csname l@#1\endcsname\relax
\typeout{** WARNING: IEEEtran.bst: No hyphenation pattern has been}%
\typeout{** loaded for the language `#1'. Using the pattern for}%
\typeout{** the default language instead.}%
\else
\language=\csname l@#1\endcsname
\fi
#2}}
\providecommand{\BIBdecl}{\relax}
\BIBdecl

\bibitem{nygaard2021real}
T.~F. Nygaard, C.~P. Martin, J.~Torresen, K.~Glette, and D.~Howard, ``Real-world embodied ai through a morphologically adaptive quadruped robot,'' \emph{Nature Machine Intelligence}, vol.~3, no.~5, pp. 410--419, 2021.

\bibitem{kang_animal_2021}
D.~Kang, S.~Zimmermann, and S.~Coros, ``Animal gaits on quadrupedal robots using motion matching and model-based control,'' in \emph{2021 IEEE/RSJ International Conference on Intelligent Robots and Systems (IROS)}.\hskip 1em plus 0.5em minus 0.4em\relax IEEE, 2021, pp. 8500--8507.

\bibitem{yang_fast_2022}
Y.~Yang, T.~Zhang, E.~Coumans, J.~Tan, and B.~Boots, ``Fast and efficient locomotion via learned gait transitions,'' in \emph{Conference on Robot Learning}.\hskip 1em plus 0.5em minus 0.4em\relax PMLR, 2022, pp. 773--783.

\bibitem{li_fastmimic_2023}
T.~Li, J.~Won, J.~Cho, S.~Ha, and A.~Rai, ``Fastmimic: Model-based motion imitation for agile, diverse and generalizable quadrupedal locomotion,'' \emph{Robotics}, vol.~12, no.~3, p.~90, 2023.

\bibitem{rathod_model_2021}
N.~Rathod, A.~Bratta, M.~Focchi, M.~Zanon, O.~Villarreal, C.~Semini, and A.~Bemporad, ``Model predictive control with environment adaptation for legged locomotion,'' \emph{IEEE Access}, vol.~9, pp. 145\,710--145\,727, 2021.

\bibitem{mnih2013playing}
V.~Mnih, K.~Kavukcuoglu, D.~Silver, A.~Graves, I.~Antonoglou, D.~Wierstra, and M.~Riedmiller, ``Playing atari with deep reinforcement learning,'' \emph{arXiv preprint arXiv:1312.5602}, 2013.

\bibitem{sutton1998reinforcement}
R.~S. Sutton, A.~G. Barto \emph{et~al.}, \emph{Reinforcement learning: An introduction}.\hskip 1em plus 0.5em minus 0.4em\relax MIT press Cambridge, 1998, vol.~1, no.~1.

\bibitem{lai_sim--real_2023}
H.~Lai, W.~Zhang, X.~He, C.~Yu, Z.~Tian, Y.~Yu, and J.~Wang, ``Sim-to-real transfer for quadrupedal locomotion via terrain transformer,'' in \emph{2023 IEEE International Conference on Robotics and Automation (ICRA)}.\hskip 1em plus 0.5em minus 0.4em\relax IEEE, 2023, pp. 5141--5147.

\bibitem{zhao_sim--real_2020}
W.~Zhao, J.~P. Queralta, and T.~Westerlund, ``Sim-to-real transfer in deep reinforcement learning for robotics: a survey,'' in \emph{2020 IEEE symposium series on computational intelligence (SSCI)}.\hskip 1em plus 0.5em minus 0.4em\relax IEEE, 2020, pp. 737--744.

\bibitem{tan_sim--real_2018}
J.~Tan, T.~Zhang, E.~Coumans, A.~Iscen, Y.~Bai, D.~Hafner, S.~Bohez, and V.~Vanhoucke, ``Sim-to-real: Learning agile locomotion for quadruped robots,'' \emph{arXiv preprint arXiv:1804.10332}, 2018.

\bibitem{smith_walk_2022}
L.~Smith, I.~Kostrikov, and S.~Levine, ``A walk in the park: Learning to walk in 20 minutes with model-free reinforcement learning,'' \emph{arXiv preprint arXiv:2208.07860}, 2022.

\bibitem{hwangbo_learning_2019}
J.~Hwangbo, J.~Lee, A.~Dosovitskiy, D.~Bellicoso, V.~Tsounis, V.~Koltun, and M.~Hutter, ``Learning agile and dynamic motor skills for legged robots,'' \emph{Science Robotics}, vol.~4, no.~26, p. eaau5872, 2019.

\bibitem{miki_learning_2022}
T.~Miki, J.~Lee, J.~Hwangbo, L.~Wellhausen, V.~Koltun, and M.~Hutter, ``Learning robust perceptive locomotion for quadrupedal robots in the wild,'' \emph{Science robotics}, vol.~7, no.~62, p. eabk2822, 2022.

\bibitem{agarwal_legged_2022}
A.~Agarwal, A.~Kumar, J.~Malik, and D.~Pathak, ``Legged locomotion in challenging terrains using egocentric vision,'' in \emph{Conference on robot learning}.\hskip 1em plus 0.5em minus 0.4em\relax PMLR, 2023, pp. 403--415.

\bibitem{ji_concurrent_2022}
G.~Ji, J.~Mun, H.~Kim, and J.~Hwangbo, ``Concurrent training of a control policy and a state estimator for dynamic and robust legged locomotion,'' \emph{IEEE Robotics and Automation Letters}, vol.~7, no.~2, pp. 4630--4637, 2022.

\bibitem{kumar_rma_2021}
A.~Kumar, Z.~Fu, D.~Pathak, and J.~Malik, ``Rma: Rapid motor adaptation for legged robots,'' \emph{arXiv preprint arXiv:2107.04034}, 2021.

\bibitem{nahrendra_dreamwaq_2023}
I.~M.~A. Nahrendra, B.~Yu, and H.~Myung, ``Dreamwaq: Learning robust quadrupedal locomotion with implicit terrain imagination via deep reinforcement learning,'' in \emph{2023 IEEE International Conference on Robotics and Automation (ICRA)}.\hskip 1em plus 0.5em minus 0.4em\relax IEEE, 2023, pp. 5078--5084.

\bibitem{fu_coupling_2022}
Z.~Fu, A.~Kumar, A.~Agarwal, H.~Qi, J.~Malik, and D.~Pathak, ``Coupling vision and proprioception for navigation of legged robots,'' in \emph{Proceedings of the IEEE/CVF Conference on Computer Vision and Pattern Recognition}, 2022, pp. 17\,273--17\,283.

\bibitem{gangapurwala_guided_2020}
S.~Gangapurwala, A.~Mitchell, and I.~Havoutis, ``Guided constrained policy optimization for dynamic quadrupedal robot locomotion,'' \emph{IEEE Robotics and Automation Letters}, vol.~5, no.~2, pp. 3642--3649, 2020.

\bibitem{bellegarda_cpg-rl_2022}
G.~Bellegarda and A.~Ijspeert, ``Cpg-rl: Learning central pattern generators for quadruped locomotion,'' \emph{IEEE Robotics and Automation Letters}, vol.~7, no.~4, pp. 12\,547--12\,554, 2022.

\bibitem{yang_multi-expert_2020}
C.~Yang, K.~Yuan, Q.~Zhu, W.~Yu, and Z.~Li, ``Multi-expert learning of adaptive legged locomotion,'' \emph{Science Robotics}, vol.~5, no.~49, p. eabb2174, 2020.

\bibitem{bellegarda_visual_2024}
G.~Bellegarda, M.~Shafiee, and A.~Ijspeert, ``Visual cpg-rl: Learning central pattern generators for visually-guided quadruped locomotion,'' in \emph{2024 IEEE International Conference on Robotics and Automation (ICRA)}.\hskip 1em plus 0.5em minus 0.4em\relax IEEE, 2024, pp. 1420--1427.

\bibitem{margolis_walk_2023}
G.~B. Margolis and P.~Agrawal, ``Walk these ways: Tuning robot control for generalization with multiplicity of behavior,'' in \emph{Conference on Robot Learning}.\hskip 1em plus 0.5em minus 0.4em\relax PMLR, 2023, pp. 22--31.

\bibitem{margolis_rapid_2022}
G.~B. Margolis, G.~Yang, K.~Paigwar, T.~Chen, and P.~Agrawal, ``Rapid locomotion via reinforcement learning,'' \emph{The International Journal of Robotics Research}, vol.~43, no.~4, pp. 572--587, 2024.

\bibitem{yang_learning_2021}
R.~Yang, M.~Zhang, N.~Hansen, H.~Xu, and X.~Wang, ``Learning vision-guided quadrupedal locomotion end-to-end with cross-modal transformers,'' \emph{arXiv preprint arXiv:2107.03996}, 2021.

\bibitem{ho_generative_nodate}
J.~Ho and S.~Ermon, ``Generative adversarial imitation learning,'' \emph{Advances in neural information processing systems}, vol.~29, 2016.

\bibitem{escontrela_adversarial_2022}
A.~Escontrela, X.~B. Peng, W.~Yu, T.~Zhang, A.~Iscen, K.~Goldberg, and P.~Abbeel, ``Adversarial motion priors make good substitutes for complex reward functions,'' in \emph{2022 IEEE/RSJ International Conference on Intelligent Robots and Systems (IROS)}.\hskip 1em plus 0.5em minus 0.4em\relax IEEE, 2022, pp. 25--32.

\bibitem{wu_learning_2023}
J.~Wu, G.~Xin, C.~Qi, and Y.~Xue, ``Learning robust and agile legged locomotion using adversarial motion priors,'' \emph{IEEE Robotics and Automation Letters}, vol.~8, no.~8, pp. 4975--4982, 2023.

\bibitem{wu_learning_2023-1}
J.~Wu, Y.~Xue, and C.~Qi, ``Learning multiple gaits within latent space for quadruped robots,'' \emph{arXiv preprint arXiv:2308.03014}, 2023.

\bibitem{kim_not_2023}
Y.~Kim, H.~Oh, J.~Lee, J.~Choi, G.~Ji, M.~Jung, D.~Youm, and J.~Hwangbo, ``Not only rewards but also constraints: Applications on legged robot locomotion,'' \emph{IEEE Transactions on Robotics}, 2024.

\bibitem{gangapurwala_rloc_2022}
S.~Gangapurwala, M.~Geisert, R.~Orsolino, M.~Fallon, and I.~Havoutis, ``Rloc: Terrain-aware legged locomotion using reinforcement learning and optimal control,'' \emph{IEEE Transactions on Robotics}, vol.~38, no.~5, pp. 2908--2927, 2022.

\bibitem{zhuang_robot_2023}
Z.~Zhuang, Z.~Fu, J.~Wang, C.~Atkeson, S.~Schwertfeger, C.~Finn, and H.~Zhao, ``Robot parkour learning,'' \emph{arXiv preprint arXiv:2309.05665}, 2023.

\bibitem{chen_learning_2020}
D.~Chen, B.~Zhou, V.~Koltun, and P.~Kr{\"a}henb{\"u}hl, ``Learning by cheating,'' in \emph{Conference on robot learning}.\hskip 1em plus 0.5em minus 0.4em\relax PMLR, 2020, pp. 66--75.

\bibitem{lee_learning_2020}
J.~Lee, J.~Hwangbo, L.~Wellhausen, V.~Koltun, and M.~Hutter, ``Learning quadrupedal locomotion over challenging terrain,'' \emph{Science robotics}, vol.~5, no.~47, p. eabc5986, 2020.

\bibitem{pinto_asymmetric_2017}
L.~Pinto, M.~Andrychowicz, P.~Welinder, W.~Zaremba, and P.~Abbeel, ``Asymmetric actor critic for image-based robot learning,'' \emph{arXiv preprint arXiv:1710.06542}, 2017.

\bibitem{yue_aacc_2022}
W.~Yue, Y.~Zhou, X.~Zhang, Y.~Hua, Z.~Wang, and G.~Kou, ``Aacc: Asymmetric actor-critic in contextual reinforcement learning,'' \emph{arXiv preprint arXiv:2208.02376}, 2022.

\bibitem{zhang_resilient_2024}
C.~Zhang, J.~Jin, J.~Frey, N.~Rudin, M.~Mattamala, C.~Cadena, and M.~Hutter, ``Resilient legged local navigation: Learning to traverse with compromised perception end-to-end,'' in \emph{2024 IEEE International Conference on Robotics and Automation (ICRA)}.\hskip 1em plus 0.5em minus 0.4em\relax IEEE, 2024, pp. 34--41.

\bibitem{schulman_proximal_2017}
J.~Schulman, F.~Wolski, P.~Dhariwal, A.~Radford, and O.~Klimov, ``Proximal policy optimization algorithms,'' \emph{arXiv preprint arXiv:1707.06347}, 2017.

\bibitem{lesort_continual_2019}
T.~Lesort, V.~Lomonaco, A.~Stoian, D.~Maltoni, D.~Filliat, and N.~D{\'\i}az-Rodr{\'\i}guez, ``Continual learning for robotics: Definition, framework, learning strategies, opportunities and challenges,'' \emph{Information fusion}, vol.~58, pp. 52--68, 2020.

\bibitem{rudin_learning_2022}
N.~Rudin, D.~Hoeller, P.~Reist, and M.~Hutter, ``Learning to walk in minutes using massively parallel deep reinforcement learning,'' in \emph{Conference on Robot Learning}.\hskip 1em plus 0.5em minus 0.4em\relax PMLR, 2022, pp. 91--100.

\bibitem{tobin_domain_2017}
J.~Tobin, R.~Fong, A.~Ray, J.~Schneider, W.~Zaremba, and P.~Abbeel, ``Domain randomization for transferring deep neural networks from simulation to the real world,'' in \emph{2017 IEEE/RSJ international conference on intelligent robots and systems (IROS)}.\hskip 1em plus 0.5em minus 0.4em\relax IEEE, 2017, pp. 23--30.

\bibitem{xie_dynamics_2021}
Z.~Xie, X.~Da, M.~Van~de Panne, B.~Babich, and A.~Garg, ``Dynamics randomization revisited: A case study for quadrupedal locomotion,'' in \emph{2021 IEEE International Conference on Robotics and Automation (ICRA)}.\hskip 1em plus 0.5em minus 0.4em\relax IEEE, 2021, pp. 4955--4961.

\bibitem{noauthor_unitree_nodate}
Unitree, ``Unitree robotics,'' \url{https://www.unitree.com/}, 2022.

\bibitem{noauthor_210810470_nodate}
V.~Makoviychuk, L.~Wawrzyniak, Y.~Guo, M.~Lu, K.~Storey, M.~Macklin, D.~Hoeller, N.~Rudin, A.~Allshire, A.~Handa \emph{et~al.}, ``Isaac gym: High performance gpu-based physics simulation for robot learning,'' \emph{arXiv preprint arXiv:2108.10470}, 2021.

\end{thebibliography}

\end{document}